\documentclass{article}

\usepackage[
    preprint,  
    nonatbib    
    ]{neurips_2020}

\usepackage[T1]{fontenc}
\usepackage[utf8]{inputenc}
\usepackage[english]{babel}

\usepackage{hyperref}
\usepackage{url}

\usepackage{booktabs}
\usepackage{amsfonts}
\usepackage{nicefrac}
\usepackage{csquotes}
\usepackage{microtype}

\usepackage{subfig}
\usepackage{graphicx}

\usepackage[backend=biber]{biblatex}
\usepackage[acronym,hyperfirst=false,nohypertypes={acronym}]{glossaries}
\usepackage{cleveref}

\graphicspath{{./figures/}}
\newacronym[shortplural={CPS}]{CPS}{CPS}{cyber-physical system}
\newacronym{AI}{AI}{Artificial Intelligence}
\newacronym{AL}{AL}{Adversarial Learning}
\newacronym{ANN}{ANN}{Artificial Neural Network}
\newacronym{ARL}{ARL}{Adversarial Resilience Learning}
\newacronym{GAN}{GAN}{Generative Adversarial Network}
\newacronym{GPGPU}{GPGPU}{General-Purpose Graphics Processing Unit}
\newacronym{DL}{DL}{Deep Learning}
\newacronym{GRU}{GRU}{Gated Recurrent Unit}
\newacronym{ICT}{ICT}{Information and Communication Technology}
\newacronym{LSTM}{LSTM}{Long-Short Term Memory}
\newacronym{AEG}{AEG}{Attack Execution Graph}
\newacronym{ML}{ML}{Machine Learning}
\newacronym{PoC}{PoC}{Proof-of-Concept}
\newacronym{RL}{RL}{Reinforcement Learning}
\newacronym{DRL}{DRL}{Deep Reinforcement Learning}
\newacronym{DNN}{DNN}{Deep Neural Network}
\newacronym{RNN}{RNN}{Recurrent Neural Network}
\newacronym{CNN}{CNN}{Convolutional Neural Network}
\newacronym{SIMD}{SIMD}{Single Instruction, Multiple Data}
\newacronym{DoE}{DoE}{design of experiments}
\newacronym{DNC}{DNC}{Differentiable Neural Computer}
\newacronym{PV}{PV}{Photovoltaic}
\newacronym{STL}{STL}{Signal Temporal Logic}
\newacronym{MTL}{MTL}{Metric Temporal Logic}
\newacronym{MITL}{MITL}{Metric Interval Temporal Logic}
\newacronym{AV}{AV}{Autonomous Vehicle}
\newacronym[shortplural={MAS}]{MAS}{MAS}{Multi Agent System}
\newacronym{RTU}{RTU}{Remote Terminal Unit}
\newacronym{TCP}{TCP}{Transmission Control Protocol}
\newacronym{LPEP}{LPEP}{Lightweight Power Exchange Protocol}
\newacronym{FMI}{FMI}{Functional Mock-Up Interface}
\newacronym{XML}{XML}{eXtensible Markup Language}
\newacronym{CTF}{CTF}{Capture The Flag}
\newacronym{A2C}{A2C}{Advantage Actor Critic}
\newacronym{A3C}{A3C}{Asynchronous Advantage Actor Critic}
\newacronym{DDPG}{DDPG}{Deep Deterministic Policy Gradient}
\newacronym{PYRATE}{PYRATE}{Polymorphic agents as cross-sectional software technology for the analysis of the operational safety of cyber-physical systems}
\newacronym{SCADA}{SCADA}{Supervisory Control and Data Acquisition}

\newglossaryentry{resilience}{
  name=resilience,
  description={The ability of a system to withstand unforseen, rare and
  potentially catastrophic events and to self-heal or improve in
  reaction to these events. Ideally resilience is increasing monotonously.}
}

\makeglossaries

\newcommand{\mA}{\ensuremath{\mathcal{A}}}  
\newcommand{\ARLBox}[2]{\ensuremath{\mathit{Box}\left\{(#1), (#2)\right\}}}
\newcommand{\ARLDiscrete}[1]{\ensuremath{\mathit{Discrete}\left\{ #1 \right\}}}

\title{The Adversarial Resilience Learning Architecture for AI-based Modelling, Exploration, and Operation of Complex Cyber-Physical Systems}

\author{%
    Eric MSP Veith\And%
    Nils Wenninghoff\\
    R\&D Division Energy\\
    OFFIS -- Insitute for Information Technology\\
    Oldenburg (Old.), Germany\\
    \texttt{firstname.lastname@offis.de}%
    \And%
    Emilie Frost%
}

\begin{document}

\maketitle

\begin{abstract}
 
    Modern algorithms in the domain of \gls{DRL} demonstrated remarkable
    successes; most widely known are those in game-based scenarios, from ATARI
    video games to Go and the StarCraft~\textsc{II} real-time strategy game.
    However, applications in the domain of modern \glspl{CPS} that take
    advantage a vast variety of \gls{DRL} algorithms are few. We assume that
    the benefits would be considerable: Modern \glspl{CPS} have become
    increasingly complex and evolved beyond traditional methods of modelling
    and analysis. At the same time, these \glspl{CPS} are confronted with an
    increasing amount of stochastic inputs, from volatile energy sources in
    power grids to broad user participation stemming from markets. Approaches
    of system modelling that use techniques from the domain of \gls{AI} do not
    focus on analysis \emph{and} operation. In this paper, we describe the
    concept of \gls{ARL} that formulates a new approach to complex environment
    checking and resilient operation: It defines two agent classes, attacker
    and defender agents. The quintessence of \gls{ARL} lies in both agents
    exploring the system and training each other without any domain knowledge.
    Here, we introduce the \gls{ARL} software architecture that allows to use
    a wide range of model-free as well as model-based \gls{DRL}-based
    algorithms, and document results of concrete experiment runs on a complex
    power grid.
  
\end{abstract}

%

\glsresetall
\section{Introduction}
\label{sec:introduction}

A \gls{CPS} is constituted of two components: On the one hand a
physical-component, in which sensors perceive the system's environment and
actuators exert actions on that environment. On the other hand a
cyber-component, where \gls{ICT} connects the distributed sensor and actuator
components of the \gls{CPS} to a computerized decision-making engine. \gls{AI}
technologies have become an essential part in almost every domain of
\glspl{CPS}. Reasons include the thrive for increased efficiency, business
model innovations, or the necessity to accommodate volatile parts of today's
critical infrastructures, such as a high share of renewable energy sources.
Over time, \gls{AI} technologies evolved from an additional input for an
otherwise solidly defined control system, through increasing the state
awareness of a \gls{CPS}, e.g. \emph{Neural State
Estimation}~\parencite{dehghanpour2018survey}, to fully decentralized but
still rule-driven systems, such as the \emph{Universal Smart Grid
Agent}~\parencite{veith2017universal}. All the way to a system where all
behavior is based on machine learning, with \emph{AlphaGo}, \emph{AlphaGo
Zero} and \emph{MuZero} probably being the most widely-known representatives
of the last
category~\parencite{silver2017mastering,schrittwieser2019mastering}.

Numerous systems are nowadays considered \glspl{CPS}, from most of today's
cars, trains, aircrafts, to, in particular, most of today's critical
infrastructures. In a recent survey we found that there is no
methodology for a comprehensive full-system
testing~\parencite{veith2019cpsanalysis}: Traditionally, \gls{CPS} analysis is
based on sound assumptions, e.g., employing models and assertions formulated
in \gls{MITL}, abstracting complexity through contracts, or employing
simulation to check whether pre-defined invariants hold. While \glspl{CPS} are
complex, the \gls{AI} domain, in particular \gls{DL} algorithms, lacks
reliable guarantees; there is no definitive way to debug a neural network.
When definite assertions cannot be given, falsifying the proposed properties
of a system is a valid tactic.  Hence, researchers are concerned with ways to
``foil'' the system, i.e., attacking it through adversarial samples or by
simply finding loopholes in its
rulesets~\parencite{Kelly2017,Teixeira2010,Ju2018b,Hirth2018}.

How much system analysis can benefit from a \gls{DRL}-based agent exploring
the system has been documented by \textcite{baker2019emergent}, where a
simulation of hide-and-seek games has uncovered bugs in the underlying
3D~engine as an unintentional side effect. Many attack vectors against
\glspl{CPS} exist, research consequently advances the hardening of \glspl{ANN}
as \gls{CPS} controllers or analyzes the behavior of \glspl{ANN} in the face
of certain activations to counter malicious inputs. But a vast research gap
exists in using \gls{AI} for model building and, subsequently, resilient
operation strategies. This gap lies, in exploring a complex \gls{CPS},
uncovering of previously undescribed or unexpected interrelation between
entities in the environment, and subsequently provide training data for a
resilient operation of the same \gls{CPS}. 

In this paper, we describe the software architecture of \gls{ARL}.
\Gls{ARL}~\parencite{Fischer2019arl} is our proposed methodology in which
independent agents explore an unknown system, either probing for weaknesses or
deriving strategies for a resilient operation. It works by two agents, an
attacker and a defender, competing against each other for control of a
\gls{CPS} model. Therefore, we begin by describing related work for
\gls{AI}-based modelling and analysis of complex systems in
\cref{sec:related-work} to describe the broader context of \gls{ARL}. In
\cref{sec:arl}, we give a brief introduction to \gls{ARL} itself, as it is a
rather young methodology. \Cref{sec:arl-software-architecture} describes the
software architecture proper, focusing on how we incorporate different
\gls{DRL} methodologies and ensure a reliable experimental process.  We then
show results of \gls{ARL} runs in a \gls{CTF}-like setting in
\cref{sec:experimental-results}. A discussion of approach and results in
\cref{sec:discussion} as well as an outlook for future development in
\cref{sec:conclusion} concludes this paper.


\section{Related Work}
\label{sec:related-work}

\gls{ARL} builds heavily on the various methodologies that have been developed
in the domain of \gls{DRL}. The first revival of \gls{DRL} research with
(end-to-end) \emph{Deep Q-Learning} is marked by the hallmark publication by
\textcite{mnih2013playing} that focused on the idea that the sequence of
observations is non-stationary and updates to the \gls{DRL} policy network are
highly correlated. As a result, the variations of Deep Q-Learning have seen
numerous advances. The ``rainbow paper'' by \textcite{Hessel2018} is a good
compilation of advances in this regard. Other approaches contributed to a
healthy community research in and applying Deep Q-Learning, such as the
\emph{Action-Branching Architectures} by \textcite{Tavakoli2017}, who address
the \emph{curse of dimensionality} experienced in Deep Q-Learning.

One reason for the \emph{curse of dimensionality} is the inability of
Q-Learning to cope with continuous action
spaces~\parencite{lillicrap2015continuous}. Policy gradient methods that
combine reward and policy directly, are architecturally more fit to cope with
continuous action spaces as they are often present in real-world scenarios.
One of the early algorithms in this regard is
REINFORCE~\parencite{Williams1992}; modern approaches are represented by the
actor-critic family, in particular \gls{A3C}~\parencite{Mnih2016},
\gls{A2C}~\parencite{mirowski2016learning}, or
ACKTR~\parencite{wu2017scalable}.

All \gls{DRL} methodologies, be it model-free ones like Q-Learning or policy
gradient algorithms, or modern model-based approaches, such as \emph{MuZero}
by \textcite{schrittwieser2019mastering}, have established a series of
benchmark-like scenarios, from the ATARI games to beating Go world champions
\parencite{silver2017mastering,Silver2017,schrittwieser2019mastering} to using
race driving simulators~\parencite{lillicrap2015continuous}.

Besides ``benchmark-like'' scenarios, \gls{DRL} already finds application in
simulations of critical infrastructures and interconnected markets. When
leaving the realm of ``pure'' \gls{AI} research towards that of \gls{CPS}
analysis and operation, the selection of \gls{DRL} methodologies becomes more
conservative. Predominantly the application of \gls{DRL} is the subject of
research, less the research on \gls{DRL} itself. Examples, specifically from
the domain of power systems, are the adaptive emergency control system by
\textcite{Huang2019}  or voltage control systems such as ``Grid Mind,''
presented by \textcite{Duan2020}: They focus on Q-Learning---often without the
advances from the Rainbow Paper---or \gls{DDPG}, as they are more readily
available from go-to libraries. \textcite{Tang2020} incorporate the idea of
two agents competing against each other in a \gls{CPS} setting; this
attack-defender scenario is an idea parallel to our
\gls{ARL}~\parencite{Fischer2019arl} concept. However, the former have
deliberately chosen a game-theoretic approach, whereas \gls{ARL} uses any
\gls{CPS} simulation without restricting itself to a formal method of
environment modelling.

However, all of these approaches treat the underlying \gls{DRL} methods as a
tool, opting to avoid a design in which algorithms, from simple Q-Learning to
complex, distributed \gls{A3C} and MuZero approaches, can be transparently
combined and even benchmarked against each other. In contrast, the well-known
\emph{OpenAI Gym} environment and its extensions bring well-known, classic or
extended settings, but have no equivalent for \gls{CPS} analysis and
operation~\parencite{brockman2016openai,10.5555/3180935,zamora2016extending,gawowicz2018ns3gym}.
Bridging the two worlds---i.e., \gls{DRL} research, and \gls{CPS} analysis and
operation research---and allowing for a transparent utilization and further
development of many advanced \gls{DL} and \gls{DRL} methodologies (including,
e.g., Meta-RL~\parencite{wang2016learning} or Neural Turing
Machines~\parencite{graves2016hybrid}), is the goal of the \gls{ARL} approach
and the subsequently developed framework we present here.


\section{The \emph{Adversarial Resilience Learning} Concept}
\label{sec:arl}

In \gls{ARL}, we define classes of agents, of which two disjoint are most
prominent: \emph{attacker agents} and \emph{defender agents}. The attacker's
goal is to de-stabilize a \gls{CPS}, the defender's utility function is based
on robost or resilient operation of that \gls{CPS}. \Gls{ARL} agents have no
knowledge of each other, which makes sense in many real-world cases, e.g., in
the power grid, where a deviation from nominal parameters can be caused by
large-scale \gls{PV} feed-in, accidents, or an actual (cyber-) attacker. As
such, agents perceive their world through the sensors they possess and act
upon their environment through actuators.

In \gls{ARL}, agents have no domain knowledge. More than that, their sensors
do not provide them with any domain-specific information. All sensors and
actuators are unlabeled; they return or accept values within a mathematical
space definition, such as \ARLDiscrete{n} for a range of discrete values \(0,
1, \dotsc, n\), or \ARLBox{l_1, l_2, \dotsc, l_n}{h_1, h_2, \dotsc, h_n} for a
bounded, n-dimensional box \([l; h] \in \mathbb{R}^n\)
\parencite{brockman2016openai}. These agents also gain rewards; the reward
function turns them into attacker and defender, or something more nuanced in
betweend, depending on the form of the function. However, the reward function
is unit-less and conveys no direct domain-specific knowledge. The general
direction of such an approach has already been verified: \textcite{Ju2018b},
for example, show that little to no topographic knowledge is necessary for an
effective attack against the power grid. The experiment results we show in
this paper also verify that no domain-specific information needs to be
conveyed at all for an effective functioning of the \gls{ARL} agents.

We note that \gls{ARL} has no connection to \gls{AL}. In \gls{AL}, the subject
of research is how to ``foil'' \glspl{ANN}, i.e., made to output widely wrong
results in the face of only minor modifications to the input. Even though
seemingly similar by name, \gls{ARL} should not be confused with \gls{AL}, as
the core problem of \gls{ARL} is not the quality of sensory inputs, but the
unknown \gls{CPS} being subject to \gls{ARL} execution. A second concept that
is potentially similar in the name only is that of \glspl{GAN}: Here, one
network, called the generator network, creates solution candidates---i.e.,
maps a vector of latent variables to the solution space---, which are then
evaluated by a second network, the
discriminator~\parencite{ghahramani2004unsupervised,goodfellow2014generative}.
Ideally, the results of the training process are results virtually
indistinguishable from the actual solution space, which is the reason
\glspl{GAN} are sometimes called ``Turing learning.'' The research focus of
\gls{ARL} is not the generation of realistic solution candidates; this is only
a potential extension of the attackers and defenders themselves. \gls{ARL},
however, describes the general concept of two agents influencing a common
model but with different sensors (inputs) and actuators (output) and without
knowing of each others presence or actions. 


\section{ARL Software Architecture}
\label{sec:arl-software-architecture}

The \gls{ARL} framework is intended to enable the training of \gls{DRL} agents
based on the \gls{ARL} concept and to evaluate environments for possible
vulnerabilities, as well as to develop strategies for a resilient operation
for these environments. In order to guarantee the domain independence of the
framework, \gls{ARL} was designed to be as modular and therefore extensible as
possible. The framework has four functional components; each component can be
individually adapted, extended or replaced.

The \gls{DoE}, as well as the setup of an experiment and the initialization of
the individual experiment runs, are combined in the \emph{experiment}
component. The \emph{agent} component embodies the \gls{DRL} agents and works
on an \emph{environment} which serves as an interface to one or more
\emph{simulations}. The encapsulation into individual components with defined
interfaces allows a separation of concerns in development and usage. This way,
the \gls{ARL} architecture can be transparently used to develop and test new
\gls{DRL} algorithms; the evaluation can then be performed on already
implemented benchmark environments.

As the goal of the \gls{ARL} methodology that is accommodated by the framework
is the analysis and operation of \glspl{CPS} systems, the experimental
environment is also a major part of the architecture. For sound and repeatable
experiments, an abstract description language is used to define an experiment
plan---including primitives for \glspl{DoE}---and setup an environment. This
includes the number and configuration of agents used as well as their
integration into the environment. An experiment generator is used to derive
concrete experiments from the experiment plan. Each experiment is then
executed. Since the number of experiments increases strongly with increasing
complexity, a decentralized execution architecture is used.
\emph{ZeroMQ}~\parencite{hintjen2013zeromq,7360036} scales \gls{ARL}
horizontally over a distributed system.

\subsection{ARL Experiment}
\label{sec:arl-software-architecture-experiment}


The \gls{ARL} architecture enables to execute comparable runs with different configuration settings, i.e., a series of soundly defined experiments. All the data required for an experiment plan are gathered in the \emph{\gls{CPS} Abtract Ontology} (\gls{CPS}-AO). This includes references to models of the \gls{CPS}, as well as to raw data for simulation, a co-simulation setup for execution, and settings for the parameters to be investigated. The \gls{CPS}-AO allows a domain-independent description of the environment in a human-understandable format. Important settings, in addition to configuration settings of the \gls{CPS} itself, are parameters for the \gls{DRL} agents, such as the strategy, the reward function, or predefined access to sensors and actuators that is given in each and every experiment run. Finally, the \gls{CPS}-AO describes sensors and actuators in the environment, but the space definitions introduced in \cref{sec:arl} are the only information attached to them.

The \gls{CPS}-AO describes only the components relevant for the experiment, but not the topology. This is already implemented in the used simulation and thus the description of an experiment remains clear. I.e., the \gls{CPS}-AO does not claim to be a universal description language for any \gls{CPS}, but a format to describe the connection of existing models, simulators, and the \gls{ARL} agent code. Hence, the term \emph{abstract} ontology. The \gls{CPS}-AO document is the source for the \emph{\gls{CPS} Experiment Generator}~(\gls{CPS}-EG) that generates concrete experiments with all necessary information to reproducibly run them.

The fan-out/fan-in type of parallelization that can be employed for multiple experiment runs is handled by the \emph{\gls{CPS} Experiment Executor} (\gls{CPS}-EE). Each individual run is handled by a \emph{governor}. Since some \gls{DRL} algorithms rely on parallel and partly asynchronous execution of worker instances, such an experiment run can consist of a multitude of simulation environments, whose handling is also the task of the governor. The handling of parallel workers, including the distribution of weight updates between the workers, is the task of an \emph{agent conductor}. It is also responsible to clone agents using \gls{DRL} strategies that do not rely on such a parallel execution, such as simple Q-learning. This way, Q-learning agents can compete with \gls{A3C} workers. Finally, the \emph{environment} class serves as control and data interface for the agents, i.e., the agents never communicate directly with a domain model, but always with a co-simulated, abstracted environment.

\begin{figure}
    \centering
    \includegraphics[width=0.8\linewidth]{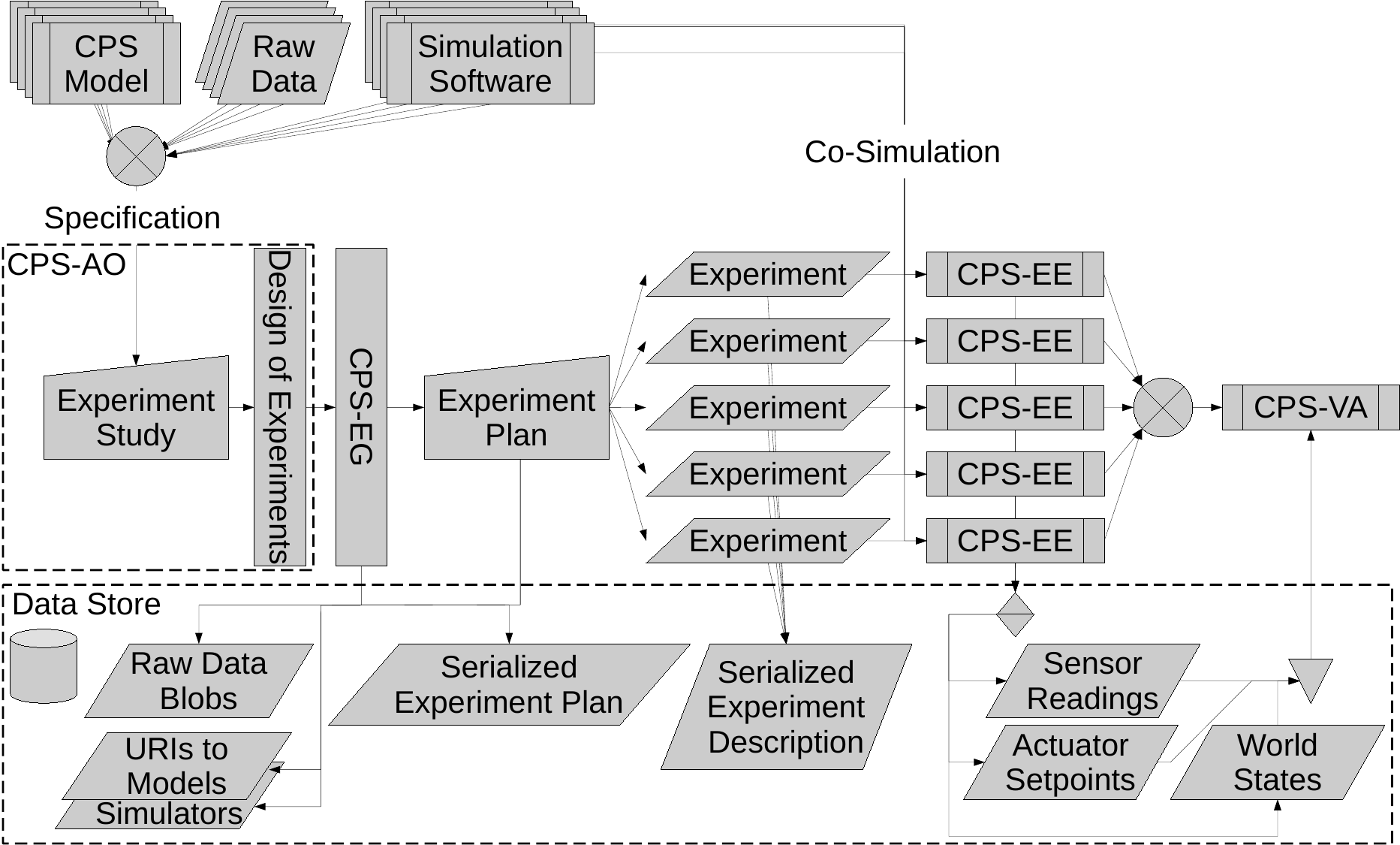}
    \caption{Experiment Flow of Execution}
    \label{fig:execution-flow}
\end{figure}

\begin{figure}[tb]
    \centering
    \includegraphics[width=\linewidth]{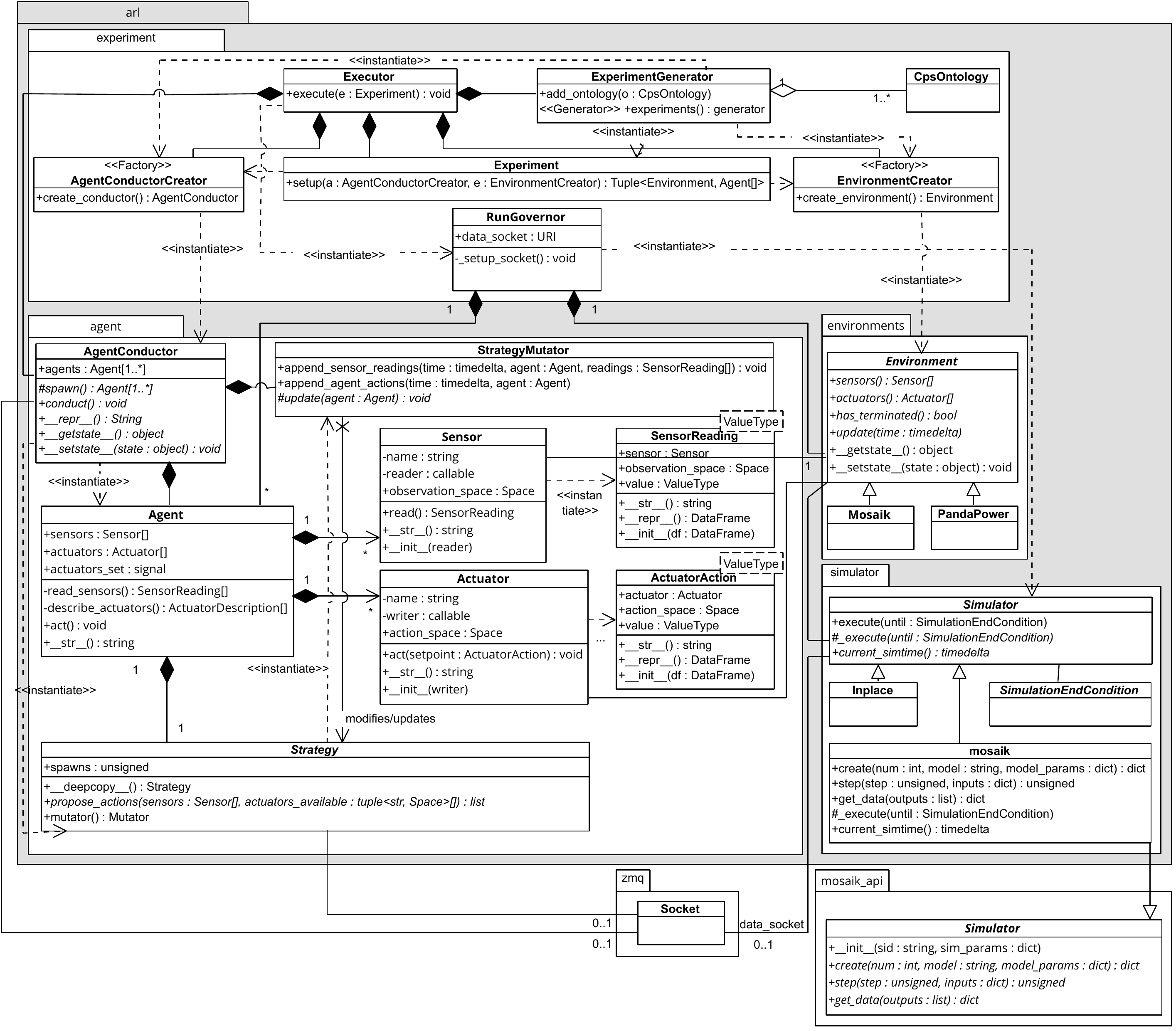}
    \caption{The Adversarial Resilience Learning Software Architecture}
    \label{fig:arl-architecture}
\end{figure}

All data from the stages of the experiment process, from raw data to software version references to experiment descriptions and parameters, to execution logs, are stored in a database as to ensure reproducibility and also allow later analysis. The whole flow of execution is depicted in \cref{fig:execution-flow}, while the more detailed class and package diagram is shown in \cref{fig:arl-architecture}.

\subsection{ARL Agents}
\label{sec:arl-software-architecture-agent}

\Gls{ARL} agents harbor the implementation of \gls{DRL} algorithms as well as other methodologies, such as neuro-evolutionary approaches \parencite{such2017deep}, or Neural Turing Machines. The overall agent is divided into a \emph{conductor} and one or more \emph{agents}/\emph{workers}. But when a new algorithm is implemented, the implementer needs to adapt only two classes: The \emph{strategy} and the \emph{strategy mutator}.

The \emph{strategy} contains the execution component of the \gls{DRL} algorithm: A strategy has one public method, called \texttt{propose\_actions(·)}. Its purpose is to map sensory inputs to actuator setpoints. For \gls{DRL} algorithms, this encapsulates \gls{ANN}, but it can as well be any simple replay, a decision tree, or any other method. The strategy also references an agent's reward function.

However, the training is implemented in the \emph{strategy mutator} as to allow asynchronous and parallel execution by more than one worker and to aid in clustering simulation approaches where machines with different hardware setups are used. For this purpose, the mutator receives both, the input values and the outputs including the rewards of all workers. It implements how a strategy's parameters  should be modified. Parameter distribution is delegated to the \emph{agent conductor}, who also implements all low-level communication facilities. This way, weight updates of the mutator are first adopted in the global network of the Agent and then distributed according to the strategy. Thus, both synchronous and asynchronous procedures are possible.

The communication between the conductor, its workers, the governor, and between agents and environment is done via message passing on a ZeroMQ bus as to decouple all modules for large-scale parallelization. The instanciation of new agents is controlled by the run governors. At the beginning of an experiment, each run governor connects one or more agents/workers with one environment instance. Once a run is complete, e.g., because the \gls{CPS} was successfully de-stabilized by an attacker, the run governor asks the conductors to spawn new workers, if necessary.

\section{Experimental Architecture Verification Setup}
\label{sec:experimental-results}

In order to verify the general feasibility of the \gls{ARL} concept and its
software architecture, we have chosen a game-like setting: A \gls{CTF}
contest. \Gls{CTF} contests have their origin in the cyber-security scene,
where two teams both control a set of servers with services they need to
defend against the respective other team, i.e., both teams need to protect
their ``bases'' as well as capture the ``flag'' from the contesting team. In
the cyber-security scene, these flags are tokens that can be read once a
service has been compromised. The \emph{DARPA Cyber Grand Challenge} was the
first to incorporate \gls{AI} into this setting~\parencite{cgc}.

To test \gls{ARL}, we chose a \emph{coin defense scenario}: The defender
starts with 10,000~points, called ``coins,'' which are being taken by the
attacker when it succeeds in bringing elements offline. The number of coins a
generator (\(c_G\) or load \(c_L\)) yields, depends on its nominal real power
characteristic \(P_N\) and the number of time steps \(t\) it remains offline
over the course of the whole simulation, denoted as \(T\):

\begin{equation}
    c_{G} = c_{L} = 0.1 P_N \frac{t}{T}~.
\end{equation}

A transformer that is being brought offline yields 20~coins, a power line
10~coins. The attacker wins when it has gained all coins from the defender
over the course of the simulation.

In our set up, we use a realistic city-state power grid. In this model, Power
is consumed by 40~load nodes (\(\cos \phi = 0.97\)), 18 of which represent
aggregated subgrids---i.e., whole districts with statistically modelled power
consumption from households, bakeries, etc.---, 22 are large-scale industry
loads. It contains 2~conventional power plants (\(\cos \phi = 0.8\)), 5~wind
farms, as well as a number of small \gls{PV} installations---mostly on
domestic rooftops---that are aggregated to 18~nodes (\(\cos \phi = 0.9\)); in
total, these nodes generate a nominal power output of 51\,MW. The grid
features a total of 22~transformers, 4 of which connect the city's medium
voltage grid to the high-voltage distribution grid; the remaining 18 connect
the city's medium voltage grid to the low-voltage grids, where households and
other regular consumers connect.

A disconnect is triggered by a violation of the grid code (Technical
Connection Rules, TCR~\parencite{ForumNetztechnikNetzbetriebimVDE.20181019})
or by constructional constraints of certain
nodes~\parencite{Brauner2012,SiemensAG.2009}. Every time an agent acts, a
power flow study is conducted and the result is checked against these
constraints. A disconnection means that the attacker has won coins; thus, from
the grid model, the coin distribution rules become obvious, as there are
multiple ways for an attacker to gain them. For example, disconnecting a
transformer means also disconnection all connected consumers, i.e., the
attacker wins not only the 20~coins associated with the transformer, but also
all coins associated with the corresponding consumers that now fall dark.
Similarly, disconnecting generators means that the city grid needs to be
supplied from the external high voltage grid; once the connecting medium-high
voltage transformer becomes overloaded, a whole district can fall dark. As
such, the \gls{ARL} attacker agent can explore many different strategies.
Similarly, different potential strategies can be explored by the defender,
too: For example, the transformer's tap changer can be used to correct the
voltage level to remain within the safe voltage band of \([0.85;
1.15]\)\,{pu}, or big loads such as industries can be scaled down, generators
used to counter fluctuating demand and supply or to control reactive power
that is needed for voltage control.

All agents have sensors that provide the current voltage level at their
respective connection point, expressed as a \ARLBox{0.85}{1.15}).
Additionally, all loads and generators `sense' their current power injection
or consumption as a value relative to their nominal input/output
(\ARLBox{0.0}{1.0}). As actuators, loads and generators provide scaling
setpoints; for \gls{DRL} algorithms that are not able to natively represent
continuous action spaces, they are discretized in 10\,\%-steps
(\ARLDiscrete{11}), otherwise, they are represented by \ARLBox{0.0}{1.0}. Tap
changers describe their possible discrete tap positions, e.g.,
\ARLDiscrete{5}. As described in \cref{sec:arl}, no sensor or actuator
contains direct domain knowledge. When agents need to assess their current
performance and world state during play and before the \gls{CTF} coins are
distributed, they use a performance function modelled to the voltage values
their sensors provide \parencite{Fischer2019arl}:

\begin{equation}
  \label{eq:reward_function}
  p_a\left(m^{(t)}\right) = {-1}^{[a\in\mA_A]}
    \exp\left[-\frac{\left(\overline{\psi_a\left(m^{(t)}\right)}-\mu\right)^2}
      {2\sigma^2}\right] 
    - c~,
\end{equation}

\noindent where \(c\), \(\mu\) and \(\sigma\) parameterize the reward curve,
\({-1}^{[a\in\mA_A]}\) negates the reward if \(a\) is an
attacker~\cite{Iverson1962}, and \(\overline{\psi_a(\cdot)}\) is the
arithmetic mean of all inputs. Note that this reward function does not include
any information specific to the energy domain. E.g., it treats the difference
between {1.0}\,{pu} and {0.8}\,{pu} similar to a reduction to {0.5}\,{pu},
even though this would mean a tremendous success to the attacker compared to a
reduction to {0.8}\,{pu}. This simplification was done deliberately as
0.8\,{pu} usually provides a general disconnection point for hardware in the
power grid, thereby already leading to a cascading blackout.

In our power grid scenario, the attacker controls all loads and static
generators, while the defender also controls all loads, static generators, and
the transformers. We have deliberately created a setting in which both agents
can influence the same elements: After all, the attacker can realistically be
a virus or botnet, which does not lock out the legitimate operator, but
overrides actions.


\section{Discussion}
\label{sec:discussion}

\begin{figure}
  \centering
  \subfloat[Defender's Coin Balance]{%
    \includegraphics[width=0.47\textwidth]{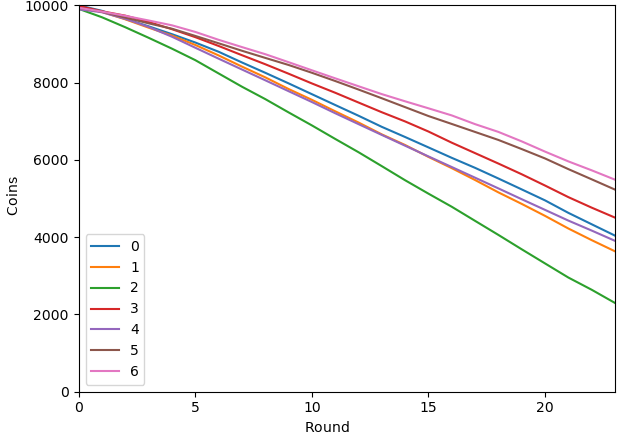}}\hfill
  \subfloat[Defender's Reward Curve]{%
    \includegraphics[width=0.47\textwidth]{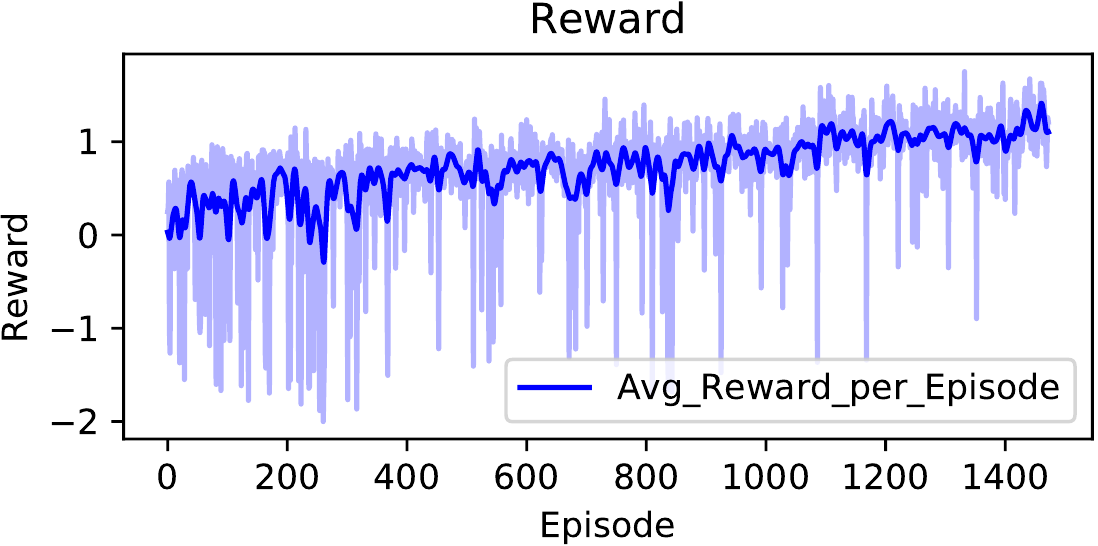}}\\
  \subfloat[Attacker Actions on Generators]{%
    \includegraphics[width=0.47\textwidth]{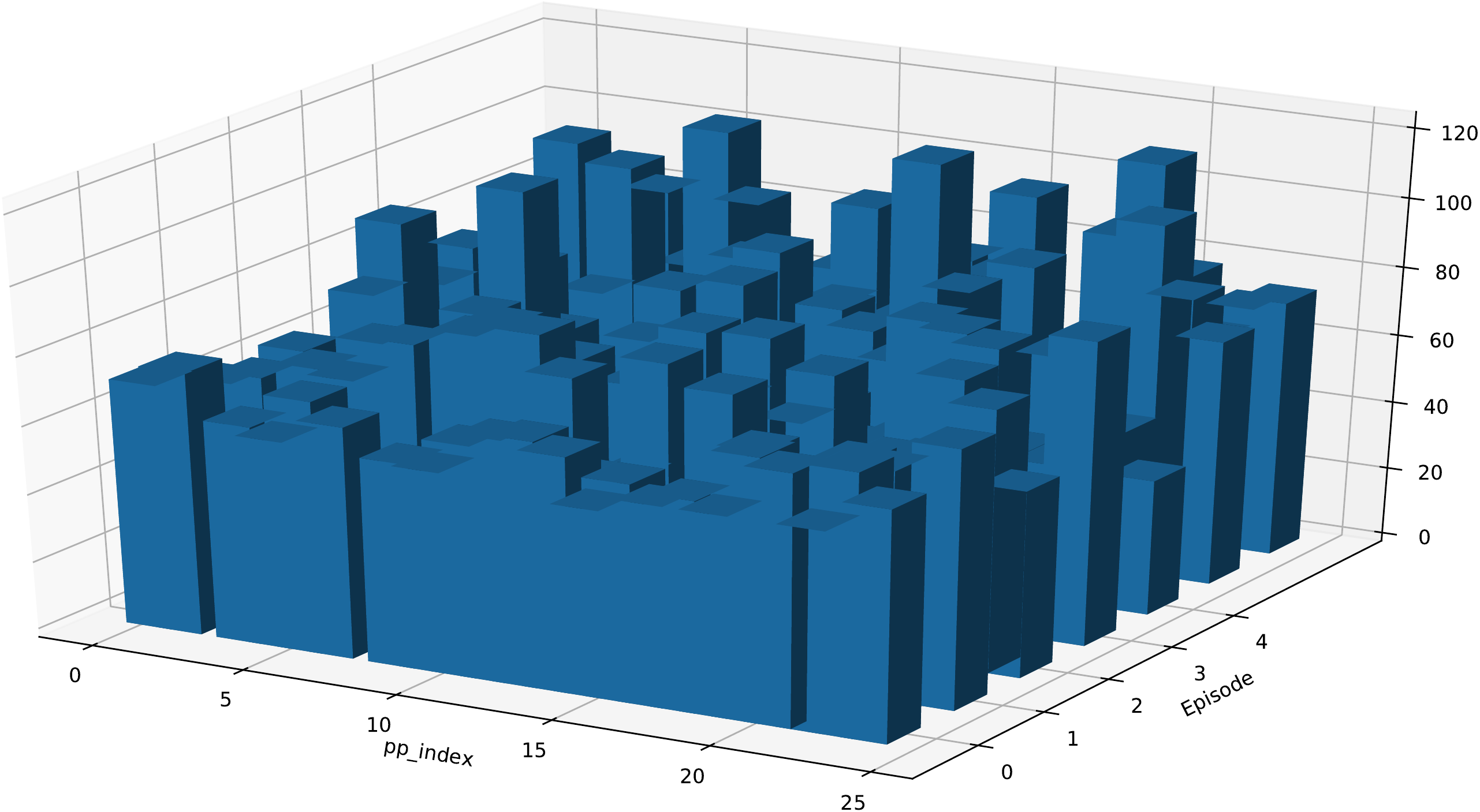}}\hfill
  \subfloat[Attacker Actions on Loads]{%
    \includegraphics[width=0.47\textwidth]{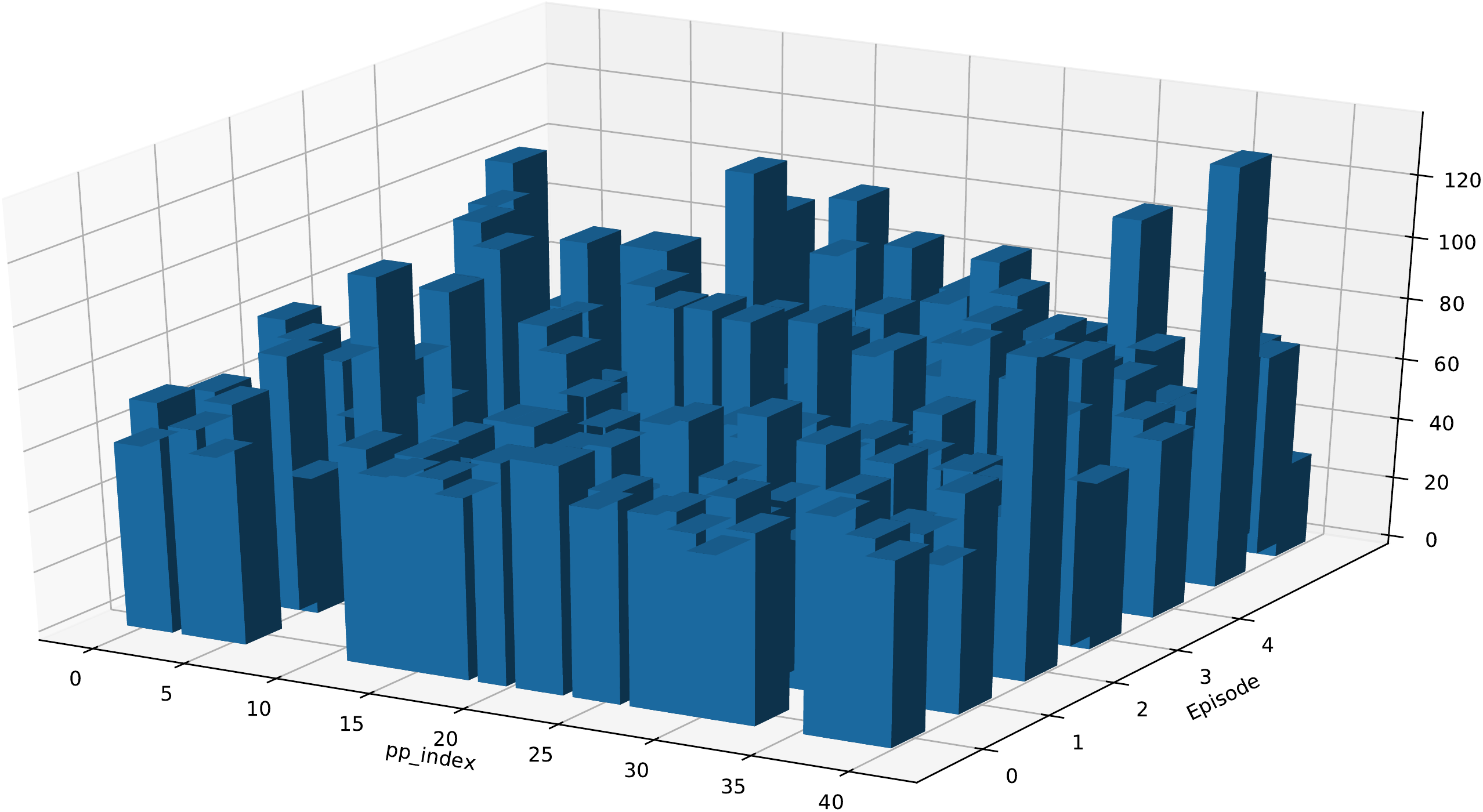}}\\
  \subfloat[Defender Actions on Generators]{%
    \includegraphics[width=0.47\textwidth]{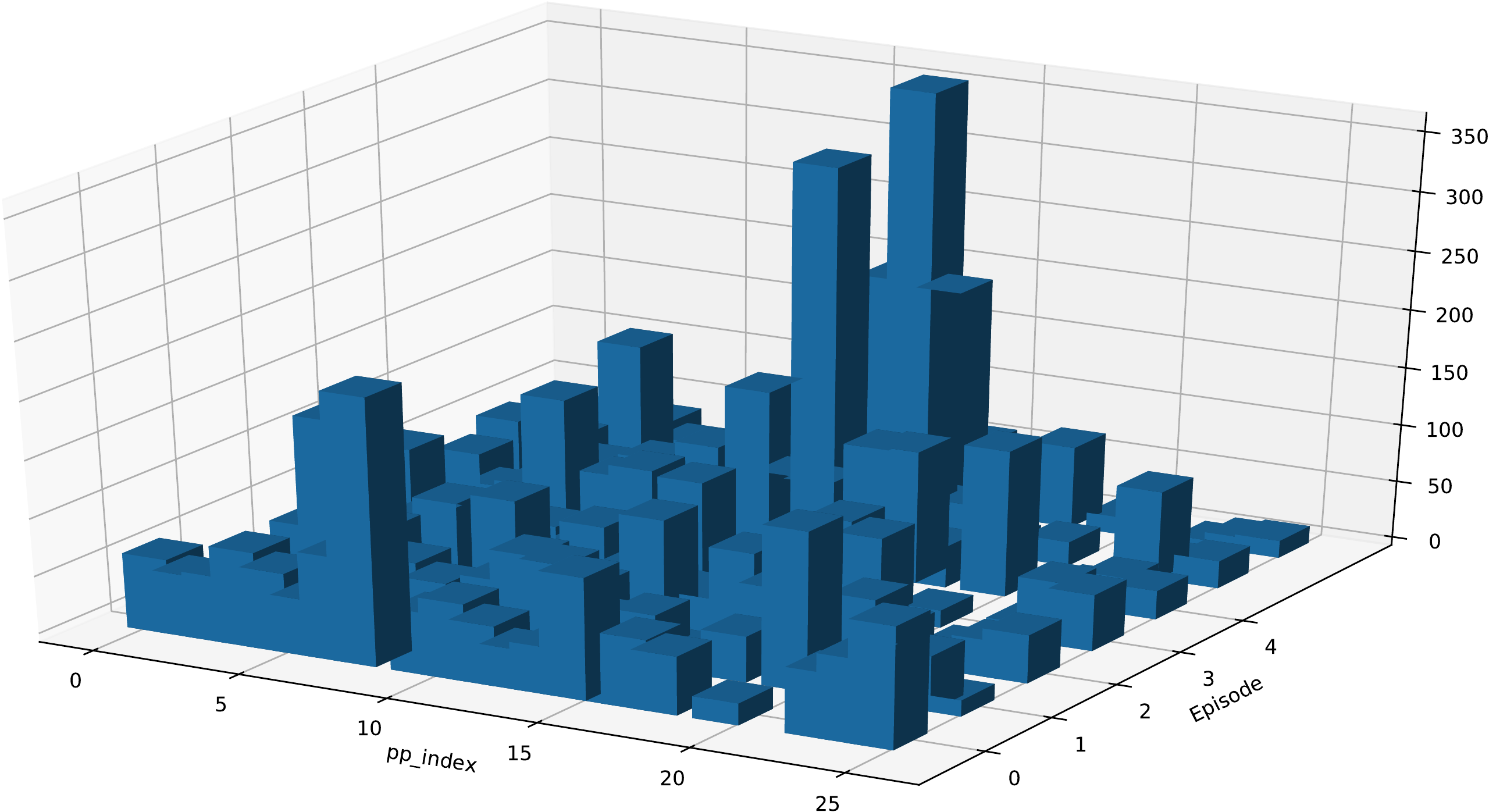}}\hfill
  \subfloat[Defender Actions on Loads]{%
    \includegraphics[width=0.47\textwidth]{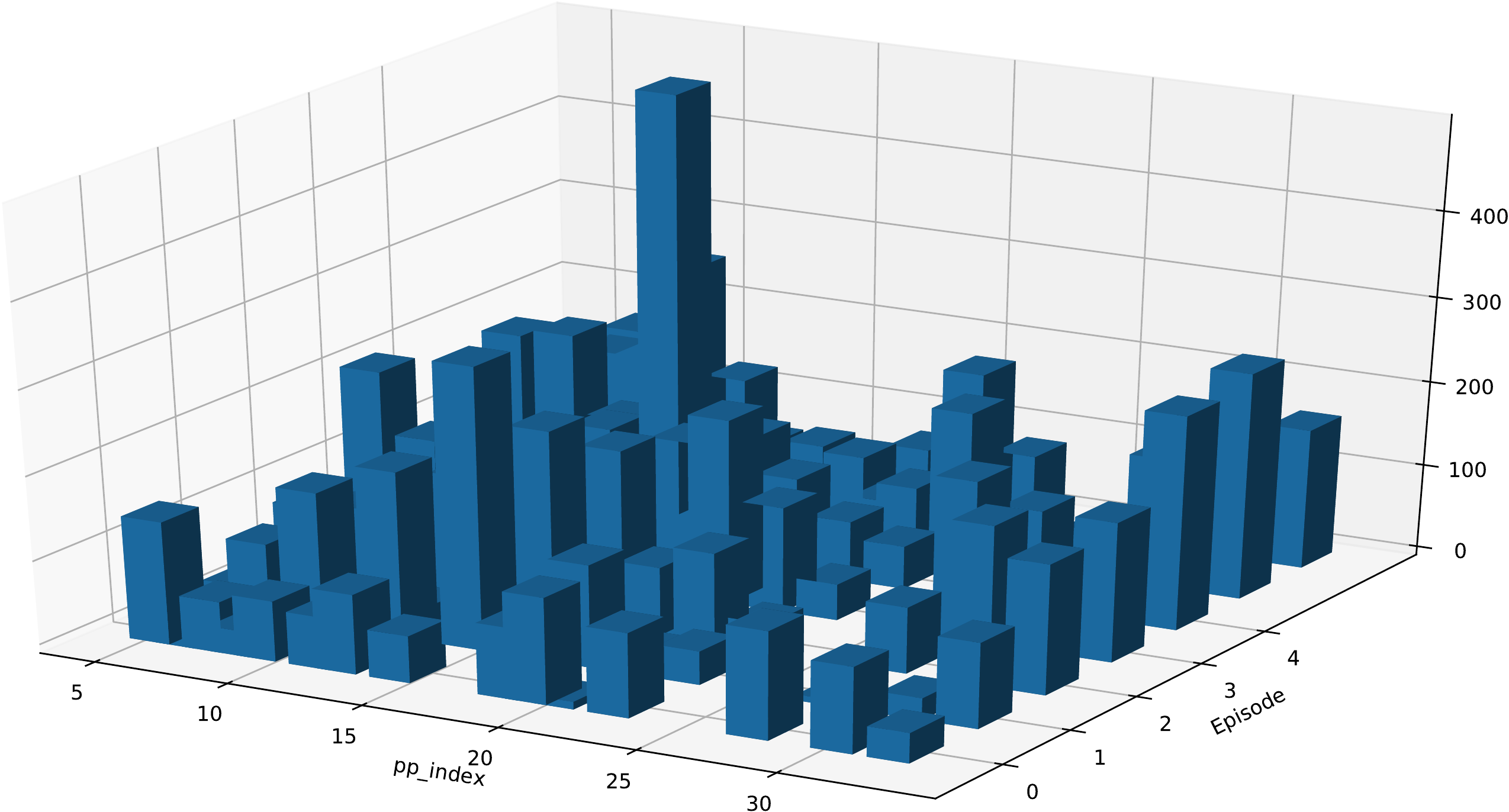}}
  \caption{Results of Experiment Runs on a Power Grid}
  \label{fig:experiment-results}
\end{figure}

We have conducted numerous \gls{CTF} tournaments---episodes organized in several
rounds---to verify our assumption that agents can learn to meaningfully attack
or operate a complex \gls{CPS}, even when interfering with each other. Also,
we wanted to verify that different algorithms can be pitched against each
other, as to verify the benchmark-notion of the \gls{ARL} architecture. Even
though this publication does not offer an extended benchmark, we can
nevertheless show the general feasibility of the \gls{ARL} approach.
\Cref{fig:experiment-results} shows averaged results for all runs.

In \cref{fig:experiment-results}(a), the attacker's success becomes apparent,
as it is able to gain coins from the defender in several \gls{CTF} rounds.
\Cref{fig:experiment-results}(b) shows the defender's reward curve that
indicates training success for the defender, i.e., the agent is able to
develop strategies to counter the attacks. Thus, each agent learns over the
episodes; interfering agents do not necessarily form a chaotic system,
which is important for \gls{ARL}-like scenarios in general.

\Cref{fig:experiment-results}(c) and~\ref{fig:experiment-results}(d)
show how much the attacker exerted control over certain actuators, summed up
and then averaged over all tournaments, with
\cref{fig:experiment-results}(c) showing generators
and~\ref{fig:experiment-results}(d) controllable loads. Similarly,
\cref{fig:experiment-results}(e) show the defender's actions on generators,
and \cref{fig:experiment-results}(f) the defender's actions on loads. This
confirms that an attacker can leverage the potential damage of each node
without requiring any topology knowledge by exploiting the other nodes'
behavior, as was shown by \textcite{Ju2018b}; also, this confirms that an
attacker can cause most difficulties for an operator by exploiting
simultaneity effects. In contrast, the defender prefers specific
generators and loads for countermeasures; this confirms what an experienced
operator would also do: Make use of system-relevant nodes to easily
redeem the power grid.

Overall, this shows that the \gls{ARL} architecture serves well to use
different \gls{DRL} algorithms to analyze and operate complex \gls{CPS}.


\section{Conclusion}
\label{sec:conclusion}\glsresetall

\Gls{ARL} is a methodology that employs \gls{DRL} algorithms to analyze and
operate complex \glspl{CPS}. We have detailed the \gls{ARL} software
architecture that allows the two \gls{ARL} agents, attacker and defender, to
work against each other in order to control the underlying \gls{CPS}. During
this, the two agents  can employ different \gls{DRL} algorithms, allowing to
exploit and analyze the characteristics of these algorithms, as well as to
compare advanced, but vastly different, \gls{DRL} approaches. We have shown
the feasibility of the \gls{ARL} approach and the architecture in a
\gls{CTF}-like tournament, with the \gls{ARL} agents competing to control a
realistic model of a complex power grid.

In the future, we plan to run a series of benchmarks on the same model,
documenting and analyzing the effect of different \gls{DRL} algorithms in the
\gls{ARL} setting, e.g., Rainbow Q-Learning against \gls{A2C} or MuZero. We
believe that the \gls{ARL} framework can be used as a complex, real-world
benchmark scenario for analysis and operation of complex critical
infrastructures and other types of \gls{CPS}, and even as a proving ground for
complex, distributed control algorithms deemed as ``robust''
\parencite{frost2020robust,veith2013lightweight,niesse2012market} for these
infrastructures. We also hope to use it as a workbench to extract strategies
for resilient operation of complex \glspl{CPS} from \gls{ARL} runs, i.e., help
to apply and advance explainable \gls{DRL}~\parencite{puiutta2020explainable}.

\newpage

\section*{Broader Impact}

The \gls{ARL} concept aims to be a methodology for \gls{AI}-based analysis and
operation of complex \gls{CPS}, specifically critical infrastructures.
\gls{ARL} explicitly `turns the tables' on the problem of non-assessible
\gls{ANN}- and \gls{DRL}-based control schemes that can be prone to
manipulation through adversarial samples, as in \gls{ARL}, the attacker is
already included. This system-of-systems-reinforcement-learning approach,
where each agent learns not only to manipulate its environment, but to do so
when another unknown party also does so, means that the agents become better
with each round they play against each other as they continuously strive to
develop better strategies.

This can be highly beneficial for operators of crictical infrastructures: The
attacker shows potential attack vectors against this infrastructure, which can
be assumed to be realistic---depending on the model---as well as
sophisticated, thanks to the defender. Furthermore, the attack scenarios can
be used as training material for personnel. The defender's use case is
obvious---i.e., operation of the \gls{CPS}---, but the domain-agnostic sensors
and actuators can yield different use cases, such as in anticipatory design,
where the operator's user interface gets modified to highlight important
information, present possible solutions, or assist in managing the flood of
system messages from \gls{SCADA} systems by priorizing or aggregating pieces
of information.

We are fully aware that the system itself could be a valuable tool for
defender \emph{and} attacker alike; it does nothing to prevent a malevolent
person from `thorwing away' the defender and utilizing the attacker on a real
piece of critical infrastructure. We hope that, the more advanced our
\gls{ARL} concept evolves to be---e.g., by incorporating neuroevolution or
similar strategies for full adaptivity---that we can also research a
explainable deep reinforcement learning technique fitting for the \gls{ARL}
concept and, based on that, a hybrid architecture with a rule-based foundation
that incorporates codified robotic laws into the \gls{ARL} agents.


\begin{ack}

This work was funded by the German Federal Ministry of Education and Research through the project \emph{PYRATE} (01IS19021A).

The authors would like to thank Sebastian Lehnhoff for his counsel and support.

\end{ack}

\printbibliography  

\end{document}